\documentclass[10pt,twocolumn,letterpaper]{article}

\usepackage{3dv}
\usepackage{times}
\usepackage{epsfig}
\usepackage{graphicx}
\usepackage{amsmath}
\usepackage{amssymb}
\usepackage{float}
\usepackage{multirow}
\usepackage{tabularx}
\usepackage{makecell}

\newcommand{\edit}[1]{#1}

\usepackage{graphicx}
\usepackage{caption}
\usepackage{subcaption}

\usepackage{algorithm}
\usepackage{algpseudocode}

\usepackage{pbox}

\usepackage[pagebackref=true,breaklinks=true,letterpaper=true,colorlinks,bookmarks=false]{hyperref}

\threedvfinalcopy 

\def\threedvPaperID{0025} 

\ifthreedvfinal\fi
\begin{document}

\title{Structure-Aware Shape Synthesis}
\author{Elena Balashova$^{1}$, Vivek Singh$^{2}$, Jiangping Wang$^{2}$, Brian Teixeira$^{2}$, Terrence Chen$^{2}$, Thomas Funkhouser$^{1}$ \\ ${}^{1}$ Princeton University, NJ, USA \\ ${}^{2}$ Siemens Healthineers, Medical Imaging Technologies, Princeton, NJ, USA}

\maketitle

\begin{abstract}
We propose a new procedure to guide training of a data-driven shape generative model using a structure-aware loss function. Complex 3D shapes often can be summarized using \edit{a coarsely defined structure} which is consistent and robust across variety of observations. However, existing synthesis techniques do not account for structure during training, and thus often generate implausible and structurally unrealistic shapes.  During training, we enforce structural constraints in order to enforce consistency and structure across the entire manifold. We propose a novel methodology for training 3D generative models that incorporates structural information into an end-to-end training pipeline.\footnote{This feature is based on research, and is not commercially available. Due to regulatory reasons its future availability cannot be guaranteed.} 
\end{abstract}

\section{Introduction}
Data-driven 3D shape synthesis is a long-standing and challenging problem in computer vision and computer graphics. The goal is to learn a model that can generate shapes within an object category suitable for novel shape synthesis, interpolation, completion, editing, and other geometric modeling applications.

\begin{figure}[t]
\centering
\includegraphics[width=0.7\linewidth]{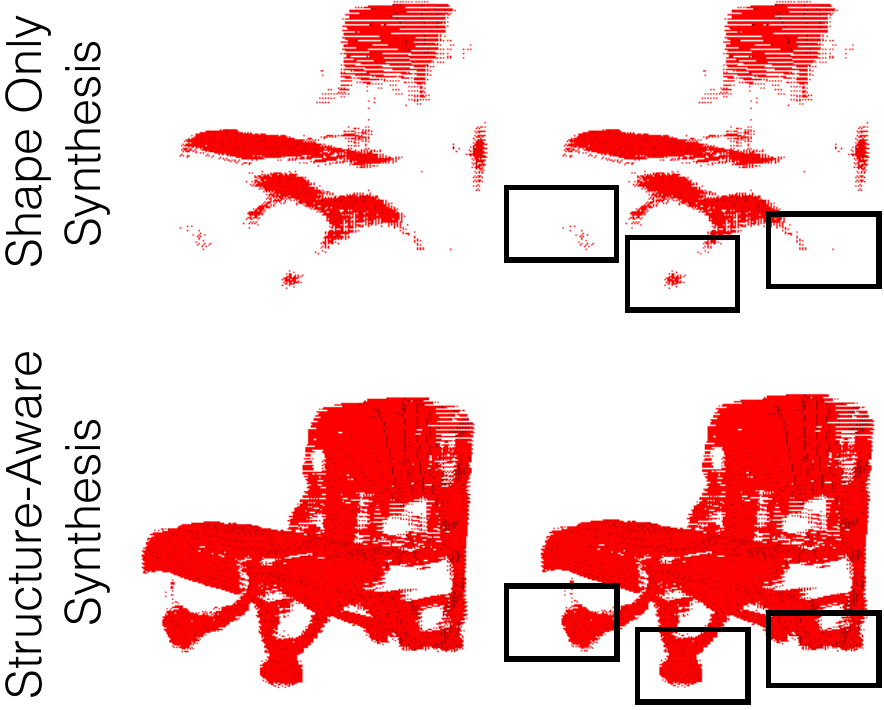}
\caption{Comparison of shape only (top row) and structure-aware synthesis (bottom row). Structure-aware shape synthesis process captures structurally meaningful shape elements, such as legs of a chair.} \label{fig:teaser}
\end{figure}

Several methods have been proposed to use a large collection of exemplar 3D shapes to train models that generate shapes using convolutional neural networks~\cite{girdhar2016learning,wu3dGAN}.   Typically, these techniques learn to generate objects from a low-dimensional vector representation to a voxel grid using supervision from synthetic CAD models (e.g., ShapeNet~\cite{chang2015shapenet}) with a loss that penalizes differences between predicted and true voxel occupancy.   Although these methods have been very successful at learning how to synthesize the coarse shapes of objects in categories with highly diverse shapes such as chairs, they have not always produced models that reconstruct important structural elements of a shape.  For example, in Figure~\ref{fig:teaser}, the reconstruction of the chair in top row has missing legs.   The problem is that the loss used for training penalizes deviations of all voxels/surfaces equally (e.g., cross entropy per voxel) without regard to preserving critical structural features.

The key idea of this paper is to learn a model that synthesizes a structural representation of an object (a set of semantic landmarks in our case) that can encourage a shape generator to reproduce critical structural elements of shape.  This idea is motivated by two observations. First, coarse-scale structure and fine-scale shape details individually convey important shape properties  -- e.g., dining chairs usually have four legs in the same general position, but with many different fine-scale shapes. Second, coarse-scale structure can often be predicted robustly even if the fine-scale shapes are not known -- e.g., predicting the positions of the feet on a dining chair does not require knowing the fine scale carvings on its surface. The novelty of our technique relies on the observation that these tasks are equally important and complementary for shape synthesis, and can be used together to learn a better generative model. We create a deep network that predicts a coarse object structure for an object (semantic landmark points) from a low-dimensional vector and then use it to define an additional loss that guides a jointly trained shape generator to reproduce the predicted coarse structure.


Our shape synthesis system is composed of three deep networks: a shape encoder, a shape generator, and a structure detector (see Figure \ref{fig:overview}). The shape encoder and generator are similar to previous work \cite{girdhar2016learning,wu3dGAN}. The key novel component is the structure detector which outputs a set of structurally important landmarks (e.g., the tip of the chair leg, the corner of the chair back etc.) used to define the aforementioned loss. This loss is used during training to encourage the shape generator to synthesize shapes with occupancy at predicted landmark locations. 

The structure detector network is pre-trained with supervised landmark points, and then the three networks are trained jointly through iterative optimization so that each can learn synergies with the others. The result is a shape generator that synthesizes shapes with more faithful structure. For example, in Figure~\ref{fig:teaser} our structure-aware shape generator (bottom row) reproduces the legs in their entirety.  Experiments using this shape generator demonstrate better results for shape interpolation and shape completion applications compared to prior work that relied on a uniform shape loss that does not consider structure.

Overall, our main research contribution is investigating how structure prediction can guide shape synthesis with deep networks. A set of landmark points is just one possible structure representation -- other possible structures include symmetries~\cite{thrun2005shape}, parts~\cite{felzenszwalb2008discriminatively} and affordances~\cite{grabner2011makes}.  We hope that this project will encourage further exploration of what types of structural properties can be predicted robustly and then used to guide shape generation with deep networks.

\section{Related Work}
 
\subsection*{Parametric Model Fitting}
Parametric model fitting methods represent a class of shapes with parametric models. Early work of active appearance models (AAM) \cite{cootes2001active} tackles registering a parametric model to images using optimization. 3D morphable models (3DMM) \cite{blanz1999morphable} align 3D scans of faces and compute a low-dimensional manifold. Similarly, statistical models of objects such as human bodies~\cite{loper2015smpl, pons2015dyna}, human hands~\cite{khamis2015learning}, and animals~\cite{chen2010inferring,zuffi20173d} have been proposed.  These methods are effective for modeling distributions in shape categories that can be easily parameterized.  However, they are difficult for complex and diverse shape categories that have multiple modes, extra and missing parts, and complex shape deformations.   Moreover, they need a consistent fit of observations to a single parametric space, which may require optimization of a non-convex objective and a robust initial guess, which is not always available, especially for partial observations as encountered in RGB-D scans.


\subsection*{Structure-Aware Shape Processing}
Previous studies have recognized the value of structure-aware shape processing, editing, and synthesis, mainly in computer graphics and geometric modeling.  In prior work, structure is broadly defined based on positions and relationships among semantic (vertical part), functional (support and stability), economic (repeatable and easy to fabricate) parts (see~\cite{mitra2013structure} for extended discussion). Symmetry-based reasoning~\cite{thrun2005shape} has been particularly well-studied in the context of shape processing~\cite{mitra2013symmetry}. Functional structure analysis, such as trained affordance models~\cite{grabner2011makes,kim2014shape2pose}, often allows for improved generalization and requires fewer examples than appearance-based feature detection. Further, co-analysis approaches, introduced by~\cite{golovinskiy2009consistent}, learn structural invariants from examples. Structure-aware approaches for shape synthesis have been suggested based on fixed relationships such as regular structures~\cite{pauly2008discovering}, partial symmetries~\cite{vst2010inverse},  probabilistic assembly-based modeling~\cite{chaudhuri2011probabilistic}, and others. Many of these structural elements are hard to annotate. In this work we focus on landmark-based structure which is relatively easy to acquire (by user annotation) and predict (for instance, using a neural network), and thus useful for data-driven shape understanding~\cite{li2017deep,wu2016single,wu20183d}. 

\subsection*{Data-Driven 3D Shape Synthesis}
Before the deep learning era, some prior works on data-driven 3D shape synthesis have proposed composing new objects from libraries of parts~\cite{funkhouser2004modeling,van2011survey}. These methods typically generate realistic shapes, but have a limited model expressiveness. More recent deep network approaches~\cite{choy20163d,girdhar2016learning} can generate higher variability of shape by learning deep generative 3D shape models on large datasets of synthetic CAD data (ex. SHAPENET~\cite{chang2015shapenet}), but generally do not optimize for structural realism of generated shapes explicitly. Variational autoencoders (VAE)~\cite{kingma2014vae} focus on reconstructing overall shapes without adding extra importance for any particular structural features.  Generative adversarial networks (GAN) focus on generating synthetic shapes that can not be distinguished from real ones by a classifier~\cite{wu3dGAN}, which ideally should detect structural irregularities, but often does not. Our focus is on improving these models with an extra loss based on structure prediction.

\section{Approach}
\begin{figure*}[t!]
\centering
\includegraphics[width=0.9\linewidth]{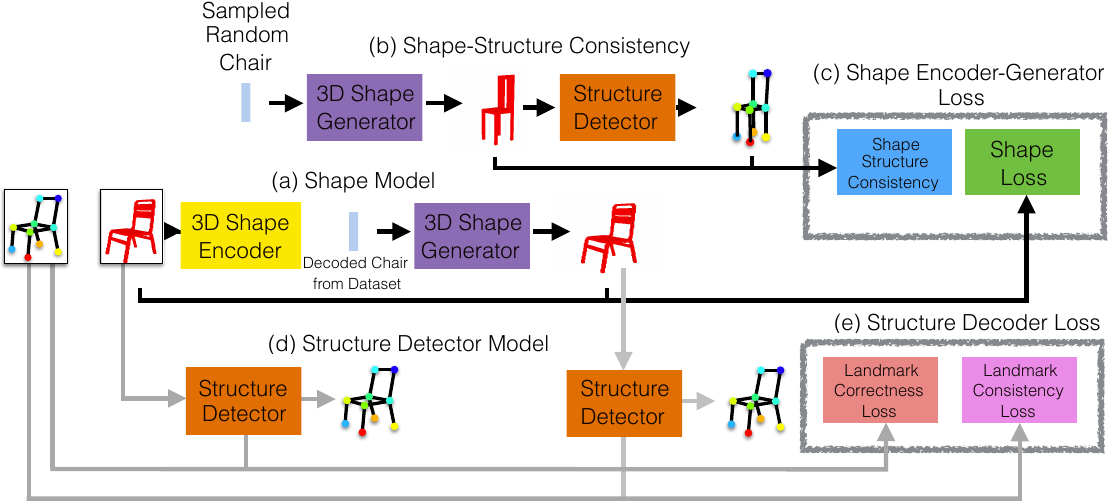}
\caption{Overview of our system, which consists of a generative shape model $(a)$ and a structure detector $(d)$. During joint training the combined shape and structure loss $(c)$ is used to train the encoder and generator and improve the generative model, while the combined landmark loss $(e)$ is propagated through the structure detector in order to predict correct structure. } 
\label{fig:overview}
\end{figure*}

An overview of our approach can be seen in Figure~\ref{fig:overview}. Our system consists of three main components: a 3D shape generator which reconstructs shapes from latent vectors (see Figure~\ref{fig:overview}$a$), a structure detector that extracts a structural representation from a 3D shape (see Figure~\ref{fig:overview}$d$), and a joint training procedure where the shape generator adjusts its predictions to better align with the structural representation output by the detector based on a new shape-structure consistency loss (see Figure~\ref{fig:overview}$b$ and Figure~\ref{fig:overview}$c$).   The key novel aspect is that the shape generator is trained to be consistent with the output of the structure detector while the structure detector is trained to be robust on the shapes produced by the generator (see Figure~\ref{fig:overview}$e$).

\edit{This structure-aware approach is very general, as it applies to any representation of structure.   In this work, for the sake of concreteness, we focus on a particular structure representation defined by a set of semantic landmark keypoints.   Our landmark-based structure for a chair, for example, represents its structurally important elements: the four tips of legs, four corners of the seat, and two top tips on the back.   We train a structure detector to predict these landmark points for any input shape, and then train the shape generator with a loss that penalizes predicted shapes that do not have mass surrounding their predicted landmarks.}

\subsection{3D Shape Encoder and Generator} \label{sec:shape}

The first component of our system is a generative model that decodes shapes from their latent representation on a manifold. Two common approaches are Variational Autoencoders (VAE)~\cite{kingma2014vae} and Generative Adversarial Networks (GAN)~\cite{goodfellow2014generative}, which are both equipped with an encoding function from the domain of observations. In this work, we focus on the first approach, which we briefly summarize here.

Define $S$ to be the set of possible shapes. Let $\Psi = \{s_1,s_2 \ldots s_m \}$ be the unlabelled set and $\Omega = \{(s_1,l_1),(s_2,l_2)\ldots (s_n,l_n)\} = (\Omega _s,\Omega _l)$ be the labelled set of examples where for each $i$, $s_i\in S$ is a 3D shape and $l_i\in \mathbb{R}^{N\times3}$ is a the set of $N$ annotated landmarks. 

Let $q_{\theta}(z | s)$ be an encoder function and let $p_{\phi} (s | z )$ be a generator function, where $s\in S$ is a shape, $z $ is its latent representation in a stochastic low-dimensional space. The loss function of the variational autoencoder is:
\begin{align}
L_{shape} = \sum _{i} \biggl[ \underbrace{-E _{z\sim q_{\theta} (z | s_i)} \left[\log p_{\phi} (s_i | z)\right]}_{\text{reconstruction loss $L_{rec}$}} +  \notag \\ 
+ \underbrace{\mathrm{KL}\left(q_{\theta} (z|s_i) \lVert  p(z)\right)}_{\text{latent distribution loss $L_{KL}$}}\biggr],
\end{align} \label{eqn:shape}
where typically $p(z) \sim \mathcal{N}(\mu,\sigma ^2)$ is enforced by minimizing Kullback Leibler (KL) divergence ($L_{KL}$).

Given that each shape $s\in S$ consists of a set of points $x_i\in \mathbb{R}^3$, the formulation of $L_{shape}(\theta,\phi)$ assumes that for each point $x_i \equiv x_ j$ for all $i,j$, i.e., all points are equally important for reconstruction. We argue that assumption is restrictive, and causes the  generative model to sometimes miss structurally important points.

Our shape encoder and generator networks are based on the architecture of Wu et al.~\cite{wu3dGAN}, where the encoder and generator mirror each other and consist of five convolutional layers each, where the encoder takes a $64\times 64 \times 64$-D voxel grid and predicts a latent $200$-D vector, while the generator predicts the voxel grid back from this latent representation.

\subsection{Structure Detector}  \label{sec:structure}

The second component of our system is a stucture detector, which takes in a 3D shape and outputs a representation of its structure (a set of landmark keypoints, in our case).  
Our structure detector network is based on Ge et al.~\cite{ge20173d}.  It consists of four convolutional layers and three fully connected layers. It takes as input a voxel grid and outputs a set of $3D$ coordinate locations for $N=10$ landmarks.

Training the structure detector is a bit tricky, since it must work for both clean meshes (like those provided with annotated structures) and synthesized ones (like those produced by the shape generator).   Ideally, we would like to obtain a structure detector $f_{\lambda} : S \rightarrow \mathbb{R}^{N x 3}$ that would predict the structure of any given shape surface of either type. Let $f_{\beta}$ be an approximation to $f_{\lambda}$. A naive way would be to use the training shapes $s_i \in \Omega _s$ and their annotations $l_i$ to train $f_{\beta}$ (structure correctness).  However, this strategy would result in a structure detector that would perform poorly on shapes with noise and other artifacts produced by the shape generator. We thus define a combined loss function that considers both correctness and robustness to these artifacts:
\begin{align}
L_{struct} &= \underbrace{\sum_ {\big\{i | s_i \in \Omega _s\big\}} \big\| l_i- f _{\beta}(s_i) \big\|}_{\text{structure correctness}}  \notag \\
			    &+ \underbrace{\sum_ {\big\{i | \hat s_i \in S \big\}} \big\| l_i- f _{\beta}(\hat s_i) \big\|}_{\text{structure robustness}}.
\end{align} \label{eqn:aff}
This strategy encourages correct structure prediction from both original and (possibly noisy) reconstructed shapes (where $\hat s_i $ is the reconstruction of training shape $s_i$).


\begin{figure}[t]
\centering
\includegraphics[width=1.0\linewidth]{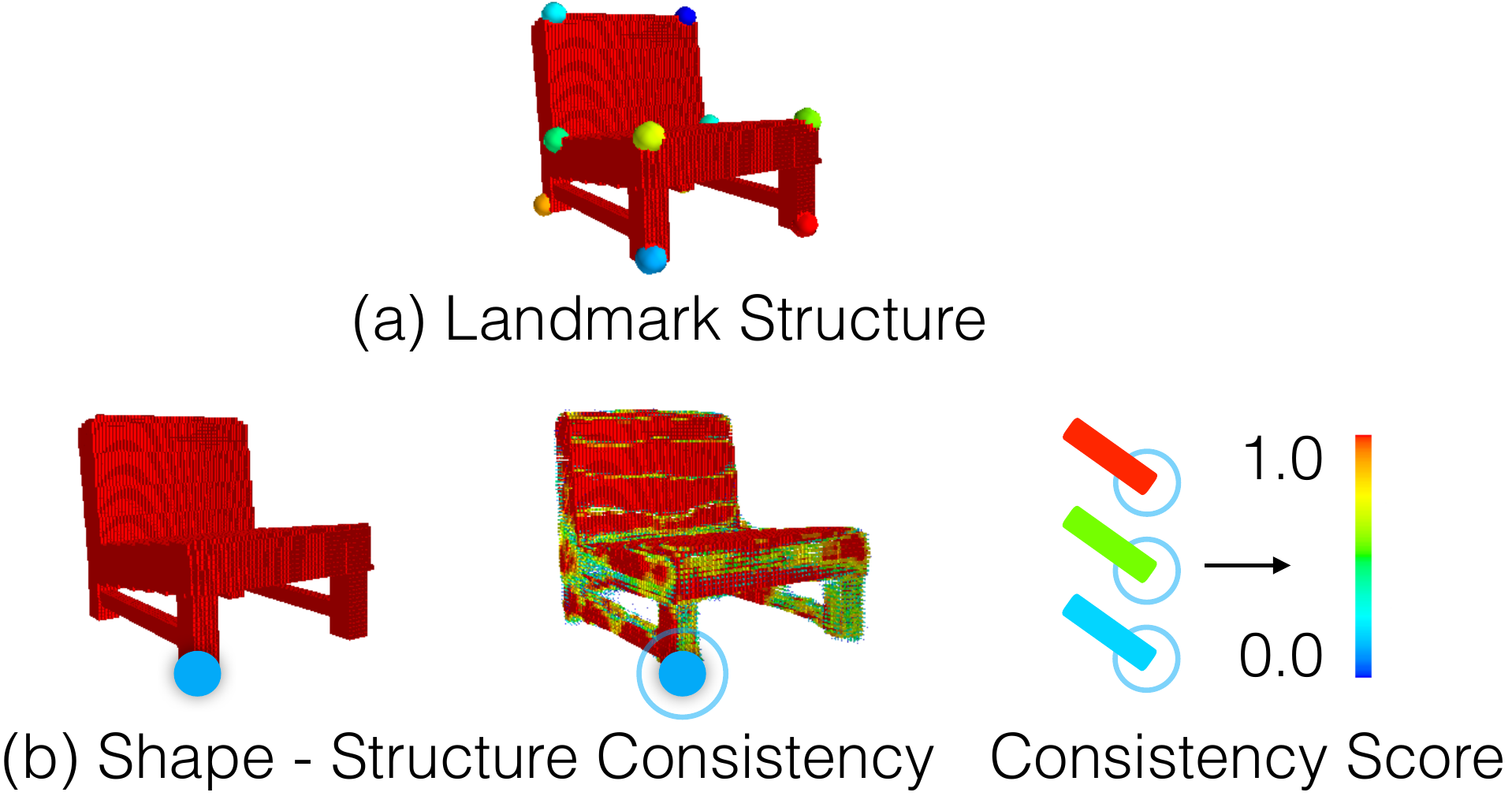}
\caption{Visualization of shape-structure consistency score. Given a coarse shape structure defined by a set of landmarks $(a)$, we measure consistency between the shape and the given structure as an additional loss on the generative model. This loss is computed by finding the highest-valued voxel in the intersection of the shape and a sphere around the landmark. The resulting shape-structure consistency loss is used at training time as an additional supervision (see Section ~\ref{sec:collab-train-sec}).} \label{fig:structure-consistency-overview}
\end{figure}

\subsection{Shape - Structure Consistency Loss}

The third and key novel component of our work is a loss function that evaluates whether a generated shape and detected structure are consistent.   
Given an example shape, we would like to measure how well does it match the structure predicted by the structure detector. Since we know apriori the structural definition of $f_{\lambda}$, we define a function of shape-structure consistency $f_{\gamma} : S \times \mathbb{R}^3 \rightarrow \mathbb{R}$ such that $f_{\gamma}(s,f_{\beta}(s)) = \|f _{\lambda}(s) - f _{\beta}(s) \|_{n_{S}}$. This function describes whether the $f _{\beta}(s)$ has correctly captured the true underlying structure $f _{\lambda}(s)$ of shape $s$, as measured by a known metric $n_S$ on shape space $S$. The shape structure consistency loss \label{eqn:aff} is defined by: 
\begin{eqnarray}
L_{consist} =  \sum_{s\in S} f_{\gamma}(s,f_{\beta}(s))
\end{eqnarray}
which encourages any point on the latent manifold to be in agreement with its defined structure. Notice that this is a stronger constraint than optimizing for shape structure consistency only on examples in the training set $s\in \Omega _s$. The latter approach may yield a sub-optimal solution where $f_{\beta}(s) = f_{\lambda}(s)$ for all $s$ in $\Omega _s$, but in general $f_{\beta}(s) \neq f_{\lambda}(s)$ for $s \in S \cap \overline{\Omega_s}$ where $S$ is the set of all possible shapes. 

We illustrate the function used for measuring shape structure consistency in Figure~\ref{fig:structure-consistency-overview}: given a shape and its predicted structure, shape structure consistency is the highest valued voxel in the vicinity of each landmark. 


\begin{algorithm}[!t]
\footnotesize
\caption{Collaborative Training Algorithm}
\label{alg:collab-train}
\begin{algorithmic}[1]
\Require{$s_i$ - ground-truth shape, $l_i$ - annotated landmarks, $\beta,\gamma,\theta,\phi$ - structure detector, shape-structure consistency estimator, shape encoder, shape generator parameters, $\lambda_1,\lambda_2$ - learning rates for structure detector and shape encoder/generator, $\mu,\sigma$ - parameters of latent distribution}
\While{$\beta,\theta,\phi$ have not converged} 
\For{each example $(s_i,l_i)\in\Omega$}
\State $z \leftarrow f_{\theta}(s_i)$ 
\State $\hat s_i \leftarrow f_{\phi}(z)$ 
\State $\hat l_i \leftarrow f_{\beta}(\hat s_i)$
\State $\dot{ l_i} \leftarrow f_{\beta}(s_i)$

\State $L_{struct}(l_i,\hat l_i,\dot{ l_i}) \leftarrow \| \hat l_i - l_i \| + \| \dot{l_i} - l_i \|$
\State $\beta \leftarrow \beta - \lambda_1 \nabla_{\beta}L_{struct}$
\EndFor

\For{each example $s_i\in\Phi$}
\State $z,\hat \mu,\hat \sigma \leftarrow f_{\theta}(s_i)$
\State $\bar{z} \leftarrow \mathcal{N}(\mu,\sigma ^2)$ 
\State $\hat s_i \leftarrow f_{\phi}(z)$ 
\State $\bar{s} \leftarrow f_{\phi}(\bar{z})$ 
\State $\bar{l} \leftarrow f_{\beta}(\bar{s})$
\State $L_{shape} \leftarrow L_{rec}(\hat s_i, s_i) + L_{KL}(\hat \mu,\hat \sigma,\mathcal{N}(\mu,\sigma ^2))$
\State $L_{consist} \leftarrow f_{\gamma}(\bar{s}, \bar{ l})$
\State $(\theta,\phi) \leftarrow (\theta,\phi) - \lambda_2 \nabla_{\theta , \phi}\big(L_{shape} + L_{consist} \big)$
\EndFor
\EndWhile
\end{algorithmic}
\end{algorithm}

\subsection{Collaborative Training Paradigm} \label{sec:collab-train-sec}

Given a pre-trained shape encoder/generator and a structure detector, we train all three models jointly. In this iterative process, the shape generator learns to respect structure, and the structure detector learns to be more robust. The training pipeline is described in Algorithm~\ref{alg:collab-train}. In this process, the structure detector receives the landmarks predicted on the ground truth and generated shapes, and compares them to the ground truth landmarks. This loss (defined by $L_{struct}$, see Section~\ref{sec:structure}) is propagated through the structure detector. The shape encoder and generator receive the ground truth shape, predict its latent representation, and output the predicted shape. Also, a random point is sampled from the manifold, and the corresponding shape and structure are predicted, and evaluated using the shape-structure consistency loss $L_{consist}$. The shape loss together with the shape-structure consistency loss are then propagated through the shape encoder/generator. 

In our experiments, we use neural networks $f_{\theta}$ ,$f_{\phi}$, $f_{\beta}$ as the encoder, generator, and structure detector functions. The shape encoder/generator and structure detector are trained separately, then the generator and structure detector are trained together with a fixed encoder, and finally the entire system is fine-tuned together.
Please see Supplementary Materials for architecture and training details.

\section{Evaluation and Discussion}
We start by evaluating our method on several tasks evaluating manifold expressiveness: interpolation and completion. We evaluate the robustness of our method when faced with sparse input (with controlled sparsity level) and check reconstruction quality on synthetic and real shapes.

\subsection{Datasets and Metrics}
We perform our experiments on the SHAPENET dataset~\cite{chang2015shapenet}, in line with prior work ~\cite{girdhar2016learning,wu3dGAN} and many others. Similar to prior work, we focus our experiments on the category of chairs, as this class of objects exhibits a challenging set of shape variability. We randomly split the dataset into $5422$ training and $1356$ testing examples for all experiments, of which $1296$ training and $374$ testing examples were annotated with landmarks by Li et al.~\cite{yi2017syncspeccnn}. For the shape completion experiment, we use the dataset of scanned models from Dai et al.~\cite{dai2017shape} that were part of our testing set. We qualitatively and quantitatively evaluate the application results using the IoU (Intersection-over-Union) metric, that was previously used for similar tasks by~\cite{choy20163d,yi2017large}.

\subsection{Shape Interpolation}
We start by qualitatively evaluating the underlying manifold learned by our system on the task of interpolation: given two topologies of chairs, we linearly interpolate their latent features in the manifold space and generate the corresponding shapes.

\subsection{Effect of Structure-Aware Learning}
\begin{figure*}[!h]
\centering
\begin{tabular}{lcc}

\multirow{2}{*}{[a]}
& \rotatebox{90}{Shape}\rotatebox{90}{Only} & \includegraphics[width=0.8\linewidth]{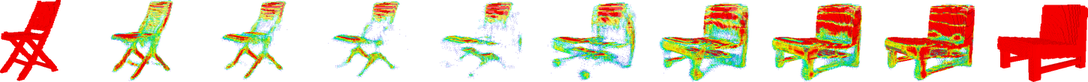} \\
& \rotatebox{90}{Struct}\rotatebox{90}{Aware} & \includegraphics[width=0.8\linewidth,trim=0 -15 0 0]{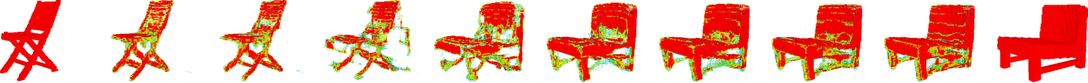} \\[0.1cm]
\hline
\\
\multirow{2}{*}{[b]}
& \rotatebox{90}{Shape}\rotatebox{90}{Only} & \includegraphics[width=0.8\linewidth]{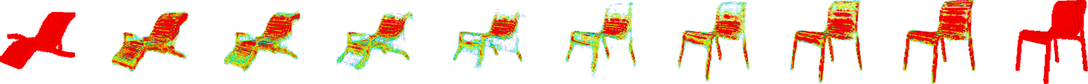} \\
& \rotatebox{90}{Struct}\rotatebox{90}{Aware} & \includegraphics[width=0.8\linewidth]{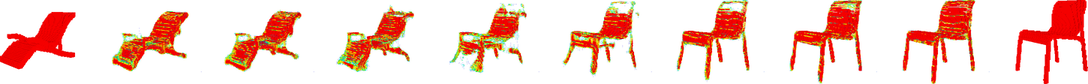} \\[0.1cm] 
\hline
\\

\multirow{2}{*}{[c]}
& \rotatebox{90}{Shape}\rotatebox{90}{Only} & \includegraphics[width=0.8\linewidth]{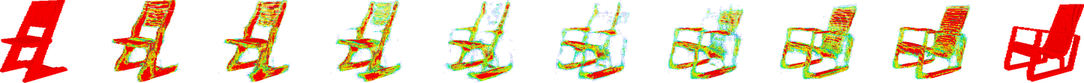} \\
& \rotatebox{90}{Struct}\rotatebox{90}{Aware} & \includegraphics[width=0.8\linewidth]{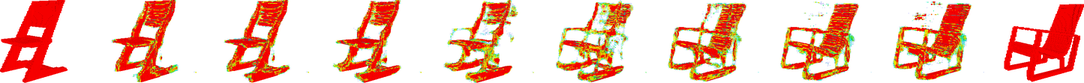} \\[0.1cm]
\hline
\\

\multirow{2}{*}{[d]}
& \rotatebox{90}{Shape}\rotatebox{90}{Only} & \includegraphics[width=0.8\linewidth]{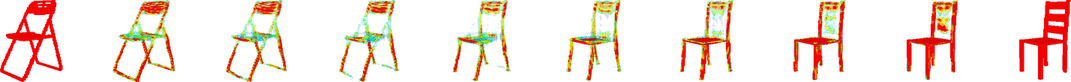} \\
& \rotatebox{90}{Struct}\rotatebox{90}{Aware} & \includegraphics[width=0.8\linewidth]{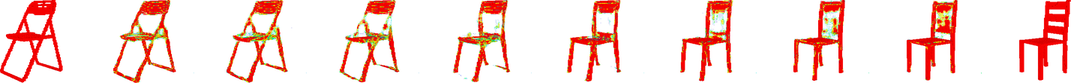} \\[0.1cm]
\hline
\\

\multirow{2}{*}{[e]}
& \rotatebox{90}{Shape}\rotatebox{90}{Only} & \includegraphics[width=0.8\linewidth]{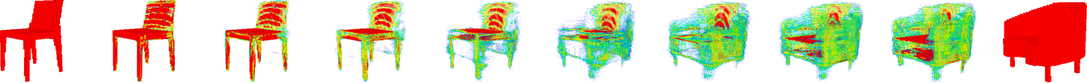} \\
& \rotatebox{90}{Struct}\rotatebox{90}{Aware} & \includegraphics[width=0.8\linewidth]{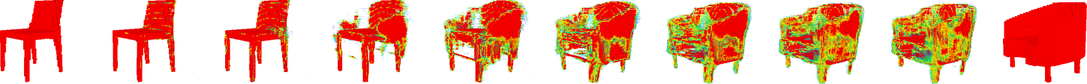} \\[0.1cm]
\hline
\\

\multirow{2}{*}{[f]}
& \rotatebox{90}{Shape}\rotatebox{90}{Only} & \includegraphics[width=0.8\linewidth]{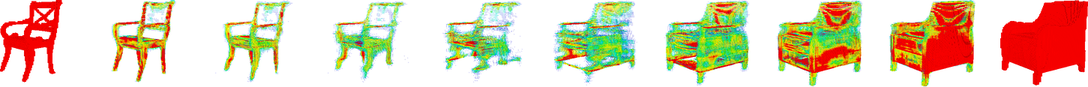} \\
& \rotatebox{90}{Struct}\rotatebox{90}{Aware} & \includegraphics[width=0.8\linewidth]{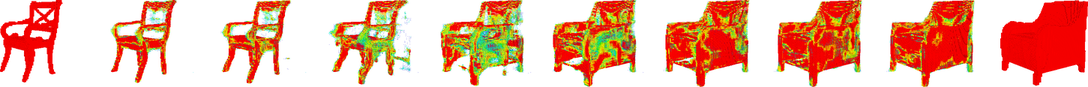} \\

\end{tabular}

\caption{Our shape generator trained with a structure detector (bottom row in each pair) produces qualitatively better interpolation of chairs than the traditional shape only method (top row in each pair).}
\label{fig:interpolation}
\end{figure*}

 We compare the interpolations of the structure-aware method \edit{(trained with shape-structure consistency)} and a baseline that does not consider structure (referred to as shape only method). The result of this experiment can be seen in Figure~\ref{fig:interpolation}. Both methods generate smooth transitions, which is expected from the underlying Gaussian manifold. Our method better captures the structure of the chair where the shape only method does not, and generally outputs higher confidence predictions (higher confidence is shown in red and lower confidence is shown in green) as compared to the baseline. The shape-aware method generates better chair legs (see examples $[a],[b],[e]$), chair backs (see examples $[a],[b],[c]$), and more confident predictions (see examples $[a]$ - $[f]$), as compared to the baseline.  

\subsection{Shape Completion on Real Scans}
We qualitatively evaluate the shape prediction on scans of real chairs. Given training data that consists of clean synthetic CAD models, reconstructing and predicting shape on real scans is challenging due to missing parts, sparsity, and reconstruction noise. We show some sample predictions of our method and compare it to the shape only baseline in Figure~\ref{fig:scan-completion}. Notice that the input scans (shown in magenta) are significantly noisier and possess missing parts, compared to their ground-truth counterparts (shown in red). For this experiment, we up-sampled the input scan point-clouds to a higher density \edit{(using binary dilation~\cite{efford2000digital})}, since the baseline method is unable to handle lower density inputs (see Section \ref{sec:sparseness}). Overall, our method tends to generate more structurally correct predictions compared to the shape only method. In the first two examples, our method correctly reconstructs a leg of the chair that was missing in the input scan. In the third example, which is missing almost a third of the chair, our method generated a complete leg, while the baseline method created a construction that is not structurally meaningful. The corresponding landmark predictions of our method are also shown. While the noise causes the structure detector to sometimes misplace landmarks from the shape by a small margin, their locations are indicative of the coarse structure in the generated shapes. We also provide quantitative results of the shape completion experiment in Table~\ref{table:completions}. Our method generates higher average IoU values on both the original and up-sampled inputs, as compared to the baseline method. In particular, the structure-aware method generates comparable prediction quality for both sampling levels, while the baseline method generates meaningful shapes only for dense inputs.

\begin{figure}[!h]
\tiny
\centering
\includegraphics[width=1.0\linewidth]{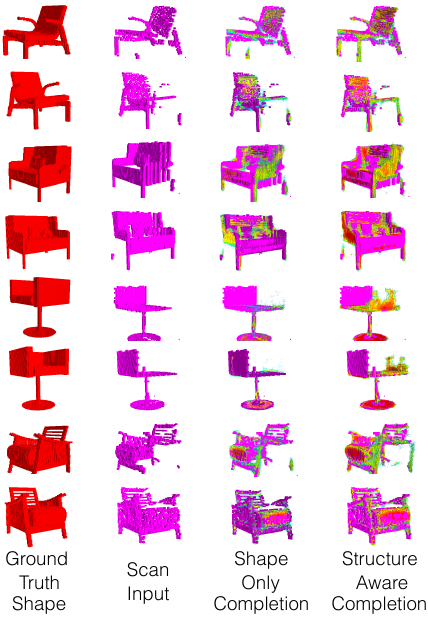}
\caption{Visualizations of the shape completion results from two different views.} \label{fig:scan-completion}
\end{figure}
\begin{table}[h]
\centering
\resizebox{\linewidth}{!}{%
\begin{tabular}{ccc}
\hline 
Metric & Shape  Only & Structure - Aware \\ \hline
Dense input  & 0.22 & \textbf{0.25} \\ 
Sparse input & 0.05 & \textbf{0.14} \\ \hline
\end{tabular}%
}
\caption{Quantitative evaluation on the scan completion task. The reported numbers are average IoU computed with respect to the ground truth shape corresponding to the scan.  Example visualizations of completions computed from dense inputs are shown in Figure~\ref{fig:scan-completion}.}
\label{table:completions}
\end{table}

\subsection{Evaluation at Different Sparseness Levels} \label{sec:sparseness}
\begin{figure}[h!]
\centering
\includegraphics[width=0.8\linewidth]{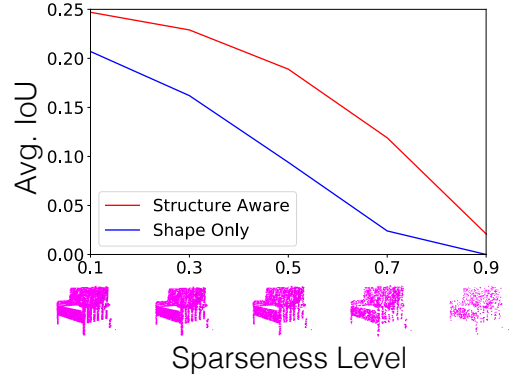}
\caption{Evaluation of scan completion at different levels of sparseness. Sparseness is controlled by dropping each point in the voxel grid with probability given by the sparseness level. The structure-aware method is more robust to sparse inputs than the shape only method.} \label{fig:sparseness-quant}
\end{figure}

We next evaluate the ability of our method to reconstruct sparse inputs. This is an important task since many real scans contain only sparse points and do not exihibit a dense surface. In Figure~\ref{fig:sparseness-quant}, we evaluate our method and the shape only method on controlled levels of sparseness. We randomly drop out points from voxel representations of scans before reconstructing them using both generative models. Here the drop-out percentage is controlled by sparseness level. We then compute the IoU of the prediction as compared to the ground truth shape corresponding to the scan. The structure-aware \edit{method} is more robust than the shape-only \edit{baseline} at all levels of dropout. This finding can be attributed to the fact that structure-aware manifold tries to map even partial inputs to complete chairs (causing an improvement in performance), while the shape only method reconstructs the input only.  

\subsection{Evaluation on Synthetic Data}
\begin{figure}[]
\centering
\includegraphics[width=0.9\linewidth]{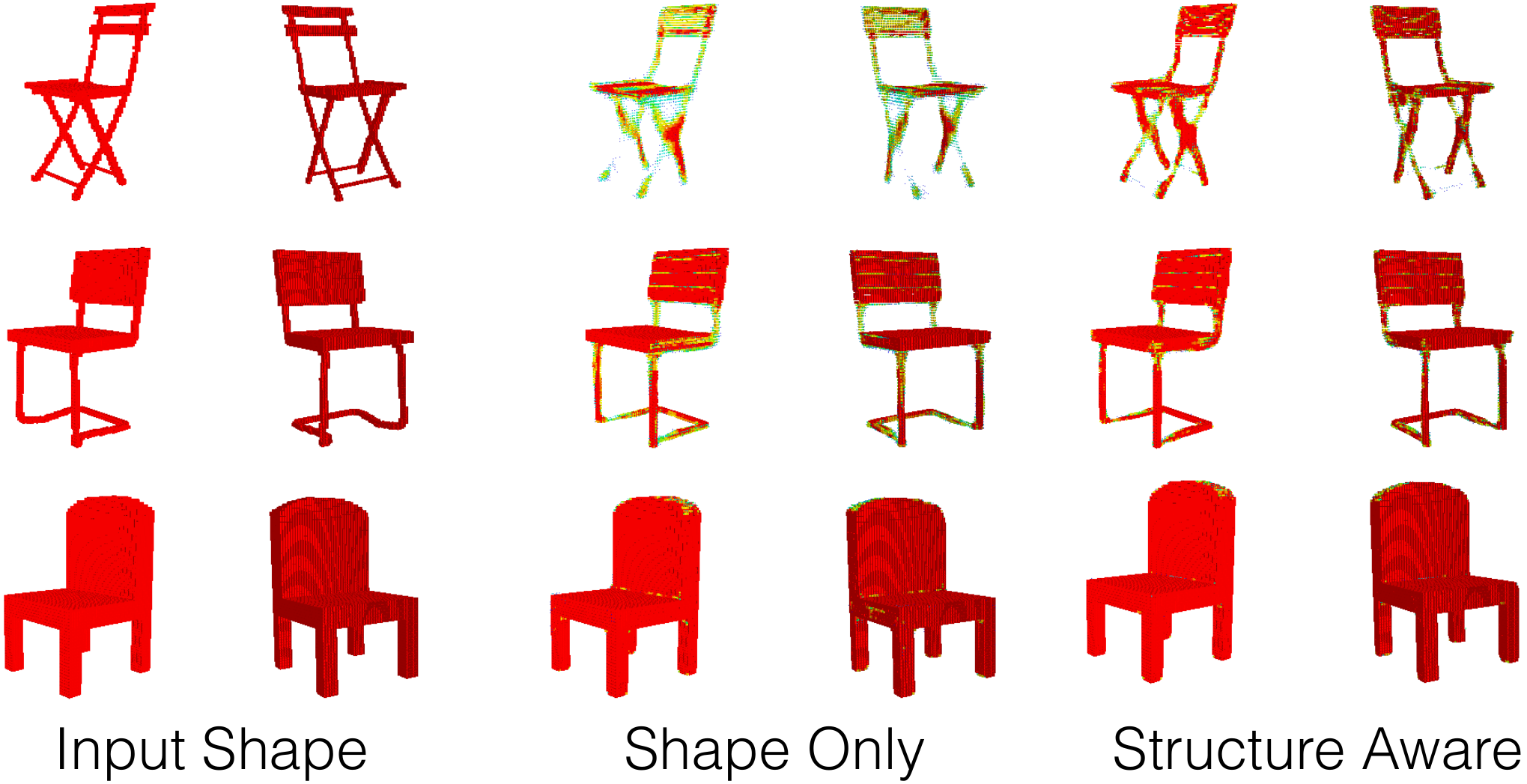}
\caption{3D reconstruction evaluation on synthetic data of the shape only and structure-aware methods. Our method achieves similar 3D shape reconstruction performance on noise-free CAD shapes. It tends to produce more confident predictions in cases of complex topology and thinner parts as it is trained to reproduce structure.} \label{fig:synthetic-reconstruction}
\end{figure}

Finally, we compare both methods on landmark prediction and 3D shape reconstruction tasks. Both the shape only and structure-aware methods are able to achieve very high quality 3D reconstruction on the test set, achieving similar performance. The performance of structure detectors (baseline trained on training shapes and our method which is additionally augmented with reconstructed data from shape generator) was also similar on the clean synthetic data, see Figure~\ref{fig:synthetic-reconstruction} for sample qualitative comparisons. However, on noisy scans originating from the generator, the baseline structure detector was unable to generate correct structure prediction, in contrast, the augmented structure detector was much more successful on this task, accurately predicting chair legs and back, see Figure~\ref{fig:synthetic-reconstruction_lm} for examples.

\setcounter{figure}{8}
\begin{figure*}[!h]
  \centering
  \begin{tabular}{ m{2cm} c }
    3D VAE GAN
    &
    \begin{minipage}{1.0\linewidth}
      \includegraphics[width=0.8\linewidth]{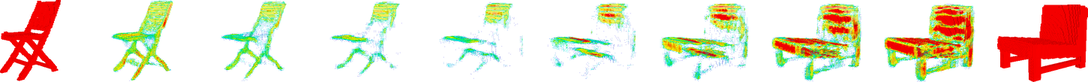}
    \end{minipage}
    \\  
    
    3D VAE 
    &
    \begin{minipage}{1.0\linewidth}
      \includegraphics[width=0.8\linewidth]{images/save_merged/chair_VAE_splitTrain_nobias_b100_aug1_merged_34.png}
    \end{minipage}
    \\ 
   
    Struct-Aware 
    &
    \begin{minipage}{1.0\linewidth}
      \includegraphics[width=0.8\linewidth]{images/save_merged/chair_joint1_lowerLR1_b2_1_iter_finetune_LM_EVALMODE_BTRAIN_merged_34.png}
    \end{minipage}
    \\ 
  \end{tabular}
\caption{Comparison between 3D VAE-GAN~\cite{wu3dGAN} (an adversarial approach), 3D VAE, and the Structure-Aware method.} \label{fig:gan_comp}
\end{figure*}

\setcounter{figure}{7}
\begin{figure}[h]
\centering
\includegraphics[width=0.8\linewidth]{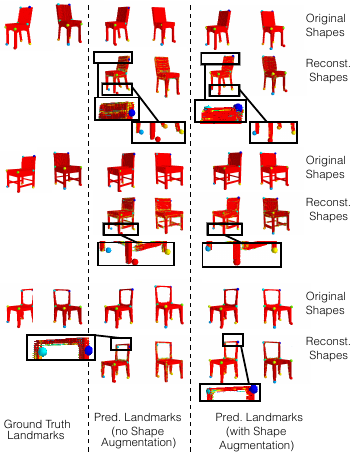}
\caption{Landmark prediction evaluation on synthetic data of structure detectors with and without data augmentation from the shape network. Both structure detectors correctly reconstruct landmark structure on shapes from the test set. The augmented structure detector is more robust to noise and predicts the correct structure on reconstructed shapes when the non-augmented detector is unable.} \label{fig:synthetic-reconstruction_lm}
\end{figure}

\section{Comparison to Adversarial Approaches}
We compare our method to an adversarial shape only baseline (3D VAE-GAN~\cite{wu3dGAN}). The result of this experiment can be seen in Figure~\ref{fig:gan_comp}. The GAN model generates slightly less noisy results (it does not produce an intermediate back part) than the VAE baseline, consistent with prior work~\cite{larsen2015autoencoding}. However, its discriminator learns the definition of realistic shapes holistically (by observing examples of correctly formed shapes), and thus does not always generate structurally meaningful outputs. 

\section{Limitations of Our Method}
Our landmark-based structure naturally constrains the types of chairs in its capacity. In this structure, there are ten landmarks, and only four landmarks allocated for legs. In the cases of chairs with five legs or no legs, such as those seen in Figure~\ref{fig:limitations}, our structure detector may still improve the shape synthesis model, but does not perform consistently since a variety of plausible landmark placements are possible for these examples (such as skipping or duplicating parts where landmarks are placed). Thus it is unclear whether the predicted structure is actually what was labelled for these examples. Additionally, the structure chosen is coarse, leading the synthesized shapes to sometimes omit finer details such as back design or additional decorative elements on chair legs (which is an inherent problem with coarse shape representations in general). These issues can be mitigated by imposing additional structural constraints (such \edit{as a different number of landmarks, importance-weighting of landmarks}, or landmark connectivity) and learning shape synthesis models at a higher resolution. \edit{Finally, our method relies on the availablility of a set of landmark-annotated examples, which may be difficult to scale up. Thus, defining and exploring automatically extracted structures is also an interesting future direction. }

\setcounter{figure}{9}
\begin{figure}[!h]
\centering
\includegraphics[width=0.65\linewidth]{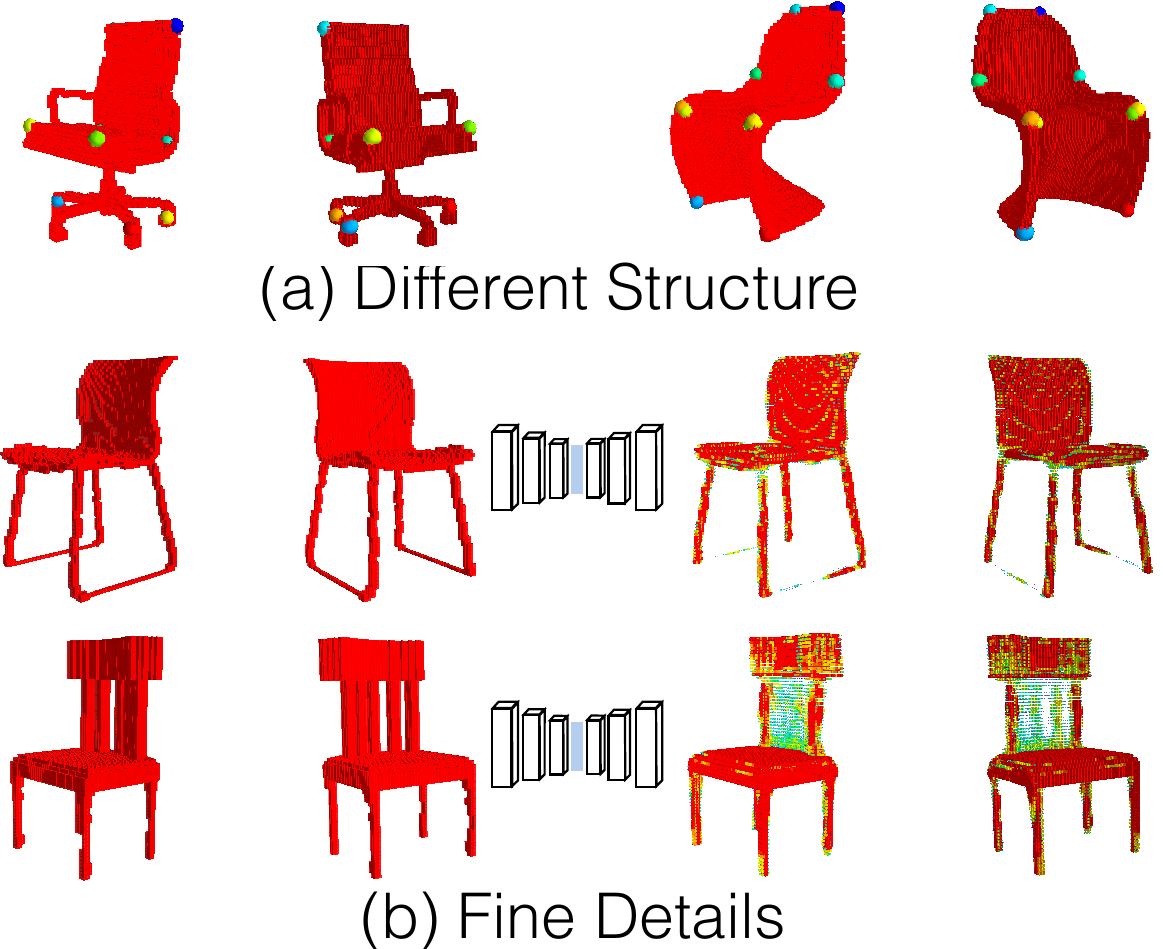} 
\caption{Our method is limited by the specific type of structure chosen. With this structure, we are only able to capture specific topologies of chairs, such as four-legged chairs. In addition, the given structure does not capture finer details, such as leg connectivity or chair back details shown above. This limitation is inherent to many coarse types of structure, and can be mitigated by introducing specific constraints, such as additional landmarks.}
 \label{fig:limitations}
\end{figure}
 \section{Conclusion and Future Work}
We present a method to train a 3D shape generation model that respects predicted structural constraints (landmark points).  By jointly training a shape generator and structure detector, the extra loss provided by shape-landmark consistency guides the generator to synthesize better shapes than without.   Though just a first step towards ``structure-aware'' shape synthesis, this approach could be applied to a number of other structural models, including skeletons or parameterized templates and in other domains, for example human bodies or faces.   It will be interesting to investigate the limits of this approach in future work.

{\small
\bibliographystyle{ieee}
\bibliography{paperbib}
}
\clearpage
\section{Supplementary Material} 
\subsection{Optimization Parameters}
\paragraph{Encoder and Generator} The shape encoder and generator networks are based on the architecture of Wu et al.~\cite{wu3dGAN}. There are five convolutional layers of output sizes $64,128,256,512,400$, with a split layer in the end to sample a   $200$-D latent vector from the produced mean and standard deviation. There are batch normalizations and rectified linear unit (ReLU) activation functions  in between all convolutional layers, and the final layer in the generator is a sigmoid function to normalize the outputs. 
\paragraph{Structure Detector} The structure detector network is based on Ge et al.~\cite{ge20173d}. It takes as input a voxel grid and outputs a set of $3D$ coordinate locations for $N=10$ landmarks. Its architecture consists of four convolutional layers with output sizes $16,32,64,128$ and kernel sizes $5,3,3,3$, with batch normalization and max-pooling layers after the first three convolutional layers and rectified linear unit (ReLU) activation functions after all four convolutions. Finally, there are three fully connected layers of output sizes $4096,1024$ and $3\dot N$ (where $N=10$ is the number of output landmarks). There are dropout layers between fully connected layers and rectified linear unit (ReLU) activation functions following all fully connected layers. 

\subsection{Training Details}
\paragraph{Data Augmentation}
To improve generalization, we randomly scale shapes in the $x,y$ and $z$ with scaling factors $s_x,s_y,s_z$ sampled uniformly in the range $[0.7,1.3)$.  The landmark positions are also updated accordingly. Data augmentation is used in all components of the system.

\paragraph{Shape Encoder and Generator}
The shape encoder/generator are trained using Adam optimizer with learning rate $0.0003$ and batch size $100$ for $200$ epochs. 

\paragraph{Structure Detector}
The structure detector is trained using Adam optimizer with learning rate $0.0001$ and batch size $32$ for $500$ epochs. 

\paragraph{Shape-Structure Consistency}
The total shape structure consistency measure consists of the shape structure consistency losses for each landmark. Given each landmark, we find all nonzero intersections of the shape voxel grid and a truncated Gaussian sphere of $\sigma=2$ around the landmark coordinate. The corresponding landmark measure is then the maximum of the element-wise product of the sphere and the shape voxel grid. The measure ($M$) is summed across all landmarks. During optimization, the loss minimized is $1/M$ to encourage increasing $M$. 

\paragraph{Collaborative Training}
During collaborative training stage, the structure detector is trained using Adam optimizer with learning rate $0.0001$ and batch size $32$. The system is trained in two stages: first, with shape encoder fixed, and shape generator and structure detector updated, second, the entire system is finetuned end-to-end. In the first stage, the shape generator is updated with Adam optimizer with learning rate $0.01$ and batch size $32$ for $150$ iterations and with learning rate $0.001$ and batch size $16,8,4,2$ for $20$ iterations to fine-tune. In the second stage,  all components are optimized at learning rate $0.000001$ and batch size $32$ for $50$ epochs.

During collaborative updates, the structure detector is optimized with equal weight for structure correctness and structure robustness losses (Section~\ref{sec:structure} in the paper). The loss for shape training is balanced with $\alpha _1=0.1$ for shape loss and $\alpha _2 =27.0$ for shape-structure consistency in order to achieve similar loss scales of both losses.   

\subsection{Additional Quantitative Experiments}
\begin{figure}[h]
\centering
\includegraphics[width=1.0\linewidth]{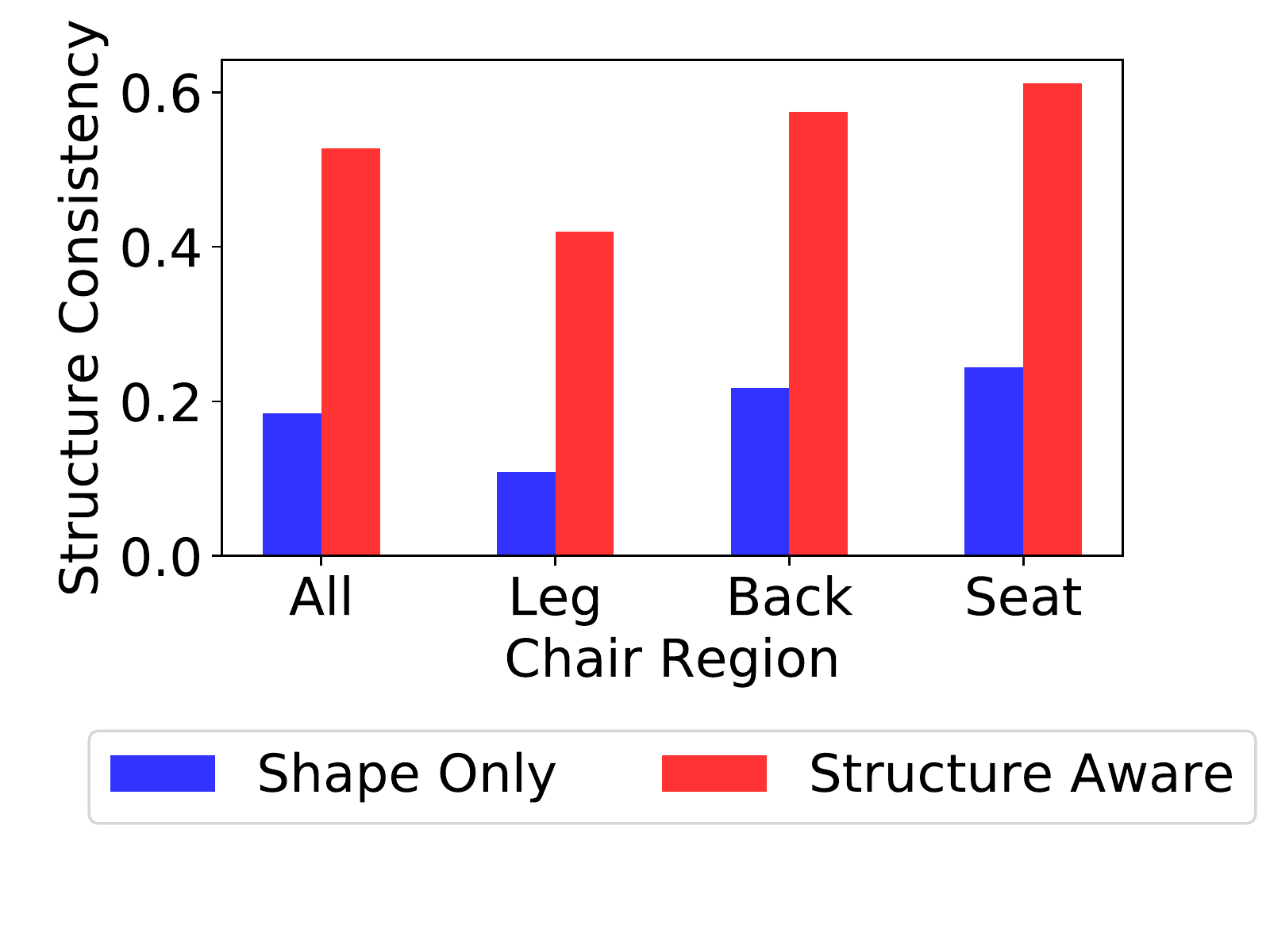}
\caption{Evaluation of avergage shape structure consistency scores by chair region. The shape structure consistency score is in range $0.0$ to $1.0$, and higher numbers indicate better performance.} \label{fig:region}
\end{figure}

We evaluate both the shape only and structure aware method using the raw shape structure consistency scores to analyze most difficult regions. In Figure~\ref{fig:region}, we provide average consistency scores per region of the chair. The structure-aware method consistently outperforms the shape only baseline. Note that for both methods, the most difficult region is the leg area, and the easiest region is the seat area. This is consistent with our initial hypothesis since chair legs exhibit much more variation than the seat area.

\begin{figure}[h]
\centering
\begin{tabular}{ccc}
\hline 
Metric & Shape  Only & Structure - Aware \\ \hline
All                        & 0.18           & 0.53 \\ 
Back (topleft)     & 0.20          &  0.57     \\ 
Back (topright)   &  0.23         &  0.58    \\ 
Leg (frontright)   &  0.09        &  0.39    \\ 
Leg (frontleft)     &   0.12        &   0.40    \\ 
Leg (backleft)     &   0.16        &   0.43  \\ 
Leg (backright)   &   0.07       &  0.46    \\ 
Seat (backleft)    &  0.26        &  0.63   \\ 
Seat (backright)  &  0.25            &  0.64    \\ 
Seat (frontleft)    &   0.23           &   0.59    \\ 
Seat (frontright)    & 0.24       & 0.59   \\ \hline 
\end{tabular}
\caption{Quantitative comparison of individual landmark shape structure consistency score.The shape structure consistency score is in range $0.0$ to $1.0$, and higher numbers indicate better performance.} \label{fig:quant-comp}

\end{figure}

\subsection{Additional Comparisons}
\begin{figure*}
\centering
\includegraphics[height=0.80\textheight]{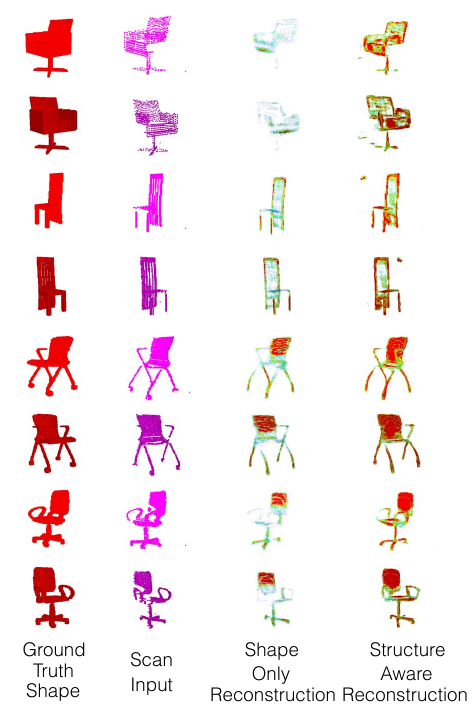}
\caption{Visualizations of reconstructions on sparse scans from two different views.} \label{fig:scan-sparse}
\end{figure*}

In Figure~\ref{fig:gan_comp}, we present additional comparisons to 3D VAE GAN and 3D VAE. In Figure~\ref{fig:scan-completion}, we show some additional completion visualizations. In Figure~\ref{fig:scan-sparse}, we show sample reconstructions on sparse scan inputs.

\begin{figure*}[!h]
  \centering
  
  \begin{tabularx}{\textwidth}{ m{2cm} c }
    3D VAE GAN
    &
    \begin{minipage}{1.0\linewidth}
      \includegraphics[width=0.8\linewidth]{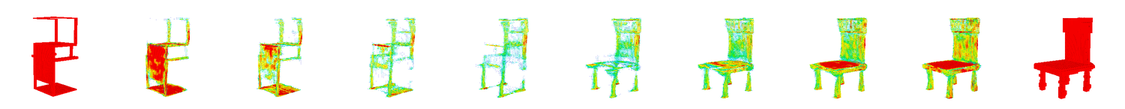}
    \end{minipage}
    \\  
    
    3D VAE 
    &
    \begin{minipage}{1.0\linewidth}
      \includegraphics[width=0.8\linewidth]{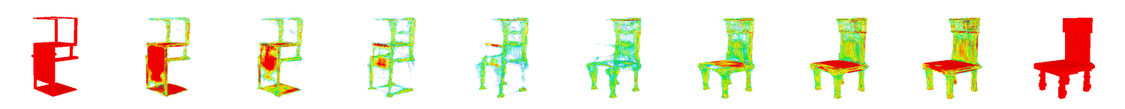}
    \end{minipage}
    \\ 
   
    Struct-Aware 
    &
    \begin{minipage}{1.0\linewidth}
      \includegraphics[width=0.8\linewidth]{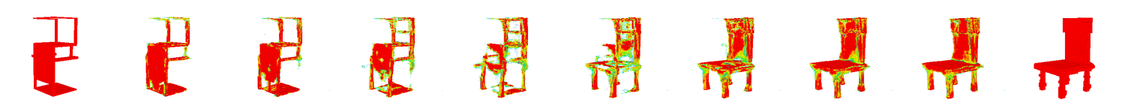}
    \end{minipage}
    \\ \hline
  \end{tabularx}
    
  \begin{tabularx}{\textwidth}{ m{2cm} c }
    3D VAE GAN
    &
    \begin{minipage}{1.0\linewidth}
      \includegraphics[width=0.8\linewidth]{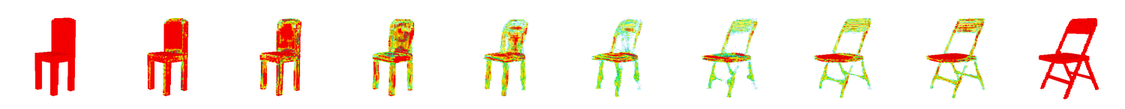}
    \end{minipage}
    \\  
    
    3D VAE 
    &
    \begin{minipage}{1.0\linewidth}
      \includegraphics[width=0.8\linewidth]{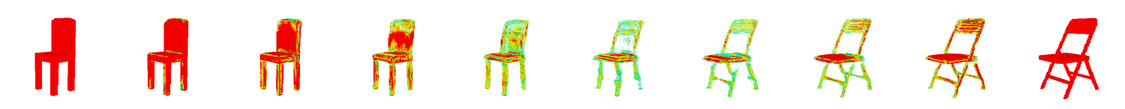}
    \end{minipage}
    \\ 
   
    Struct-Aware 
    &
    \begin{minipage}{1.0\linewidth}
      \includegraphics[width=0.8\linewidth]{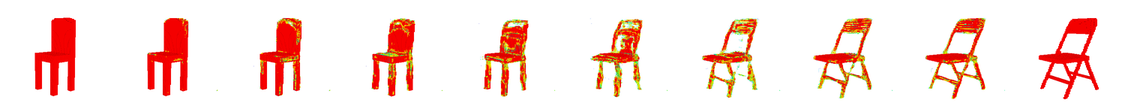}
    \end{minipage}
    \\ \hline
  \end{tabularx}

  \begin{tabularx}{\textwidth}{ m{2cm} c }
    3D VAE GAN
    &
    \begin{minipage}{1.0\linewidth}
      \includegraphics[width=0.8\linewidth]{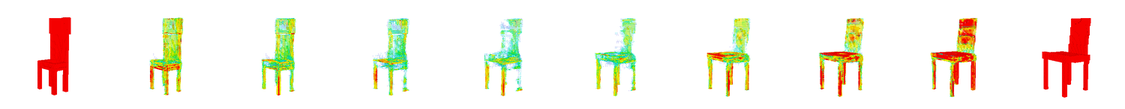}
    \end{minipage}
    \\  
    
    3D VAE 
    &
    \begin{minipage}{1.0\linewidth}
      \includegraphics[width=0.8\linewidth]{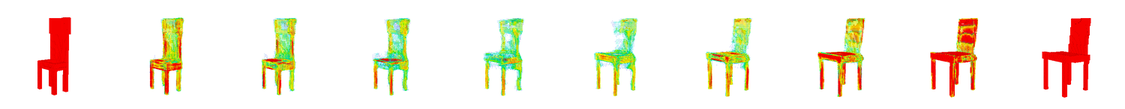}
    \end{minipage}
    \\ 
   
    Struct-Aware 
    &
    \begin{minipage}{1.0\linewidth}
      \includegraphics[width=0.8\linewidth]{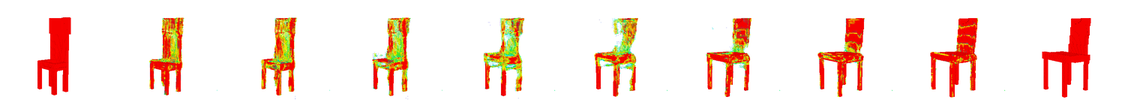}
    \end{minipage}
    \\ \hline
  \end{tabularx}
  \begin{tabularx}{\textwidth}{ m{2cm} c }
    3D VAE GAN
    &
    \begin{minipage}{1.0\linewidth}
      \includegraphics[width=0.8\linewidth]{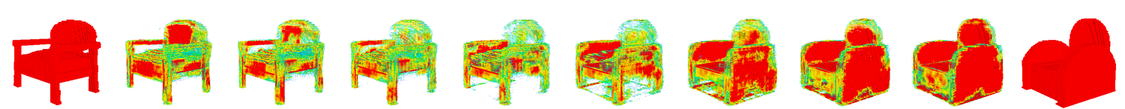}
    \end{minipage}
    \\  
    
    3D VAE 
    &
    \begin{minipage}{1.0\linewidth}
      \includegraphics[width=0.8\linewidth]{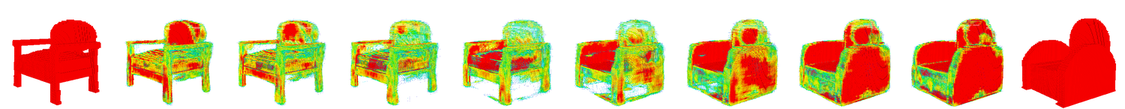}
    \end{minipage}
    \\ 
   
    Struct-Aware 
    &
    \begin{minipage}{1.0\linewidth}
      \includegraphics[width=0.8\linewidth]{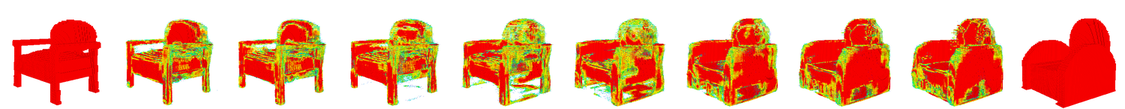}
    \end{minipage}
    \\ \hline
  \end{tabularx}
  
  \begin{tabularx}{\textwidth}{ m{2cm} c }
    3D VAE GAN
    &
    \begin{minipage}{1.0\linewidth}
      \includegraphics[width=0.8\linewidth]{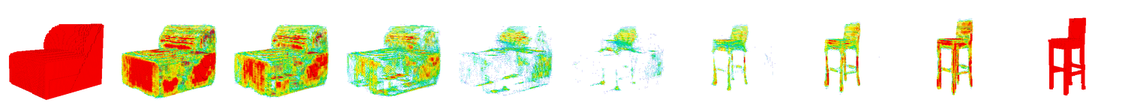}
    \end{minipage}
    \\  
    
    3D VAE 
    &
    \begin{minipage}{1.0\linewidth}
      \includegraphics[width=0.8\linewidth]{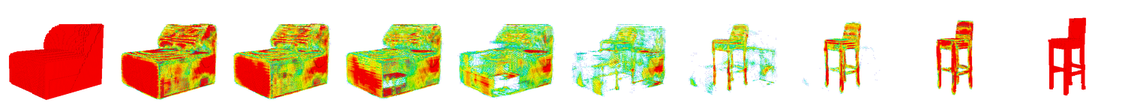}
    \end{minipage}
    \\ 
   
    Struct-Aware 
    &
    \begin{minipage}{1.0\linewidth}
      \includegraphics[width=0.8\linewidth]{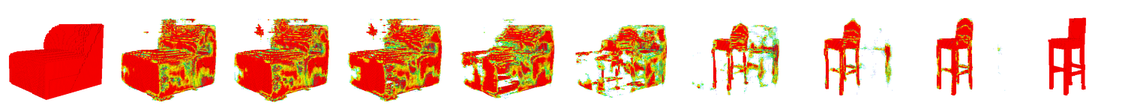}
    \end{minipage}
    \\ \hline
  \end{tabularx}

\caption{Additional comparisons between 3D VAE-GAN~\cite{wu3dGAN} (an adversarial approach), 3D VAE, and the Structure-Aware method.} \label{fig:gan_comp}
\end{figure*}

\begin{figure*}
  \ContinuedFloat 
  \begin{tabularx}{\textwidth}{ m{2cm} c }
    3D VAE GAN
    &
    \begin{minipage}{1.0\linewidth}
      \includegraphics[width=0.8\linewidth]{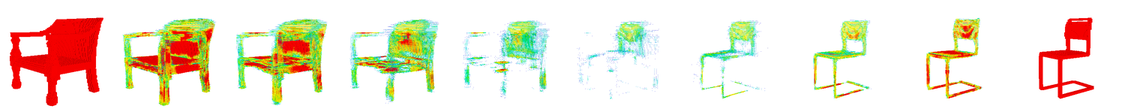}
    \end{minipage}
    \\  
    
    3D VAE 
    &
    \begin{minipage}{1.0\linewidth}
      \includegraphics[width=0.8\linewidth]{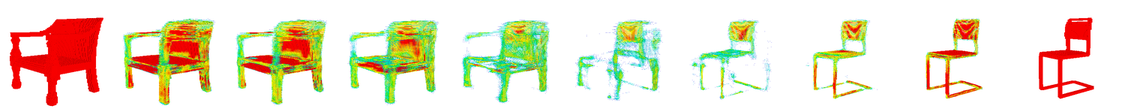}
    \end{minipage}
    \\ 
   
    Struct-Aware 
    &
    \begin{minipage}{1.0\linewidth}
      \includegraphics[width=0.8\linewidth]{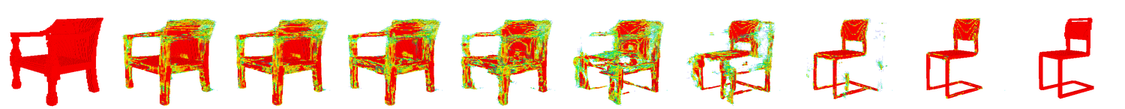}
    \end{minipage}
    \\ \hline
  \end{tabularx}
  
  \begin{tabularx}{\textwidth}{ m{2cm} c }
    3D VAE GAN
    &
    \begin{minipage}{1.0\linewidth}
      \includegraphics[width=0.8\linewidth]{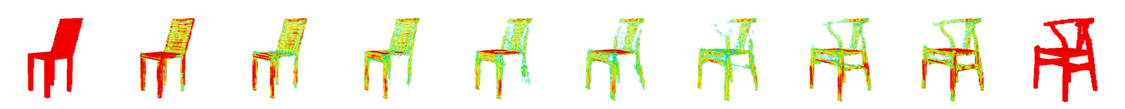}
    \end{minipage}
    \\  
    
    3D VAE 
    &
    \begin{minipage}{1.0\linewidth}
      \includegraphics[width=0.8\linewidth]{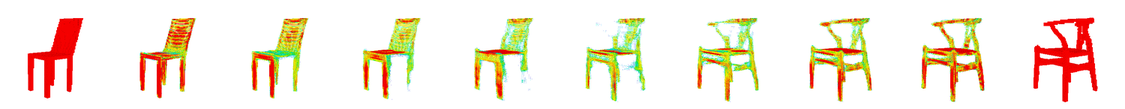}
    \end{minipage}
    \\ 
   
    Struct-Aware 
    &
    \begin{minipage}{1.0\linewidth}
      \includegraphics[width=0.8\linewidth]{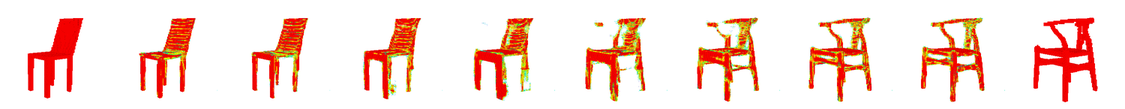}
    \end{minipage}
    \\ \hline
  \end{tabularx}

  \begin{tabularx}{\textwidth}{ m{2cm} c }
    3D VAE GAN
    &
    \begin{minipage}{1.0\linewidth}
      \includegraphics[width=0.8\linewidth]{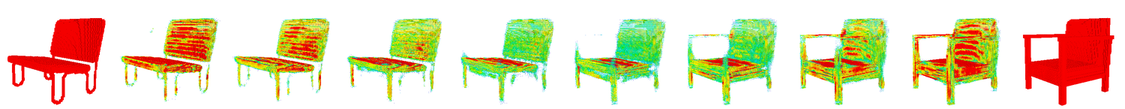}
    \end{minipage}
    \\  
    
    3D VAE 
    &
    \begin{minipage}{1.0\linewidth}
      \includegraphics[width=0.8\linewidth]{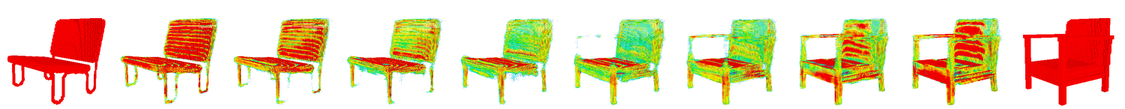}
    \end{minipage}
    \\ 
   
    Struct-Aware 
    &
    \begin{minipage}{1.0\linewidth}
      \includegraphics[width=0.8\linewidth]{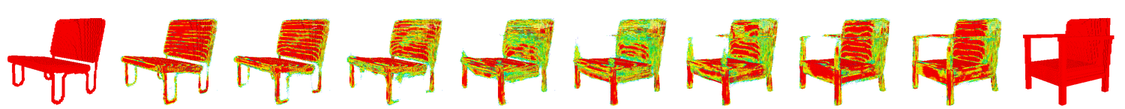}
    \end{minipage}
    \\ \hline
  \end{tabularx}
  
  \begin{tabularx}{\textwidth}{ m{2cm} c }
    3D VAE GAN
    &
    \begin{minipage}{1.0\linewidth}
      \includegraphics[width=0.8\linewidth]{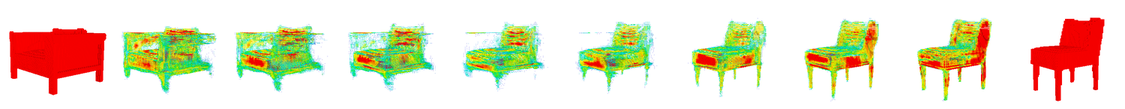}
    \end{minipage}
    \\  
    
    3D VAE 
    &
    \begin{minipage}{1.0\linewidth}
      \includegraphics[width=0.8\linewidth]{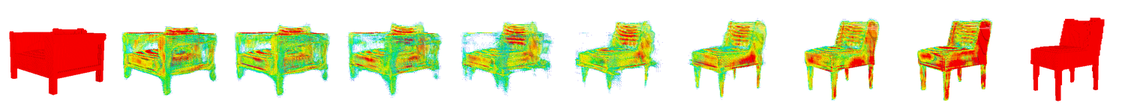}
    \end{minipage}
    \\ 
   
    Struct-Aware 
    &
    \begin{minipage}{1.0\linewidth}
      \includegraphics[width=0.8\linewidth]{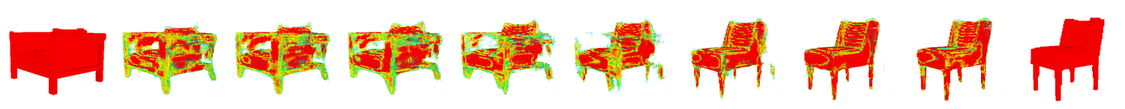}
    \end{minipage}
    \\ \hline
  \end{tabularx}
  
  \begin{tabularx}{\textwidth}{ m{2cm} c }
    3D VAE GAN
    &
    \begin{minipage}{1.0\linewidth}
      \includegraphics[width=0.8\linewidth]{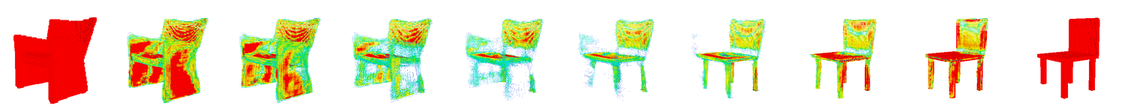}
    \end{minipage}
    \\  
    
    3D VAE 
    &
    \begin{minipage}{1.0\linewidth}
      \includegraphics[width=0.8\linewidth]{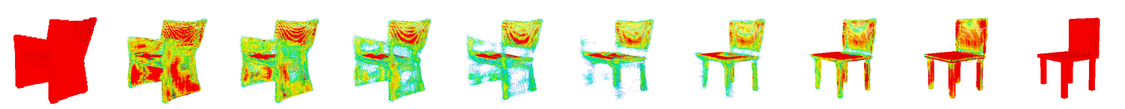}
    \end{minipage}
    \\ 
   
    Struct-Aware 
    &
    \begin{minipage}{1.0\linewidth}
      \includegraphics[width=0.8\linewidth]{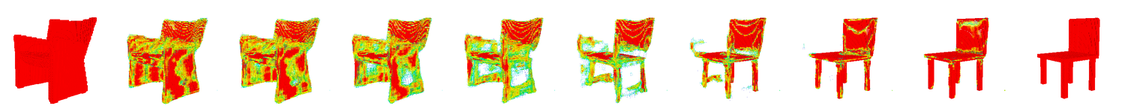}
    \end{minipage}
    \\ \hline
  \end{tabularx}

\caption{Additional comparisons between 3D VAE-GAN~\cite{wu3dGAN} (an adversarial approach), 3D VAE, and the Structure-Aware method.} \label{fig:gan_comp}
\end{figure*}

\begin{figure*}[!t]
  \ContinuedFloat 
  \begin{tabularx}{\textwidth}{ m{2cm} c }
    3D VAE GAN
    &
    \begin{minipage}{1.0\linewidth}
      \includegraphics[width=0.8\linewidth]{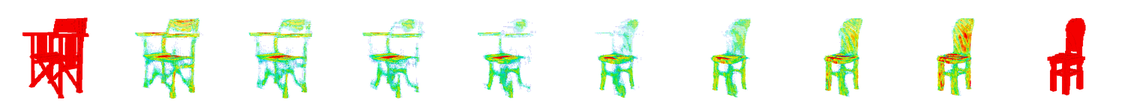}
    \end{minipage}
    \\  
    
    3D VAE 
    &
    \begin{minipage}{1.0\linewidth}
      \includegraphics[width=0.8\linewidth]{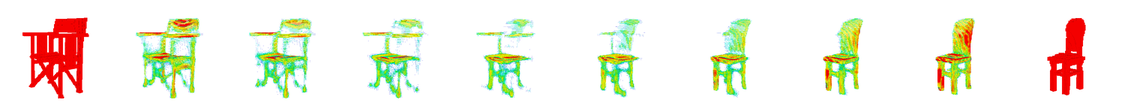}
    \end{minipage}
    \\ 
   
    Struct-Aware 
    &
    \begin{minipage}{1.0\linewidth}
      \includegraphics[width=0.8\linewidth]{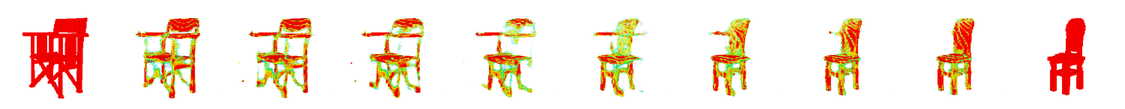}
    \end{minipage}
    \\ \hline
  \end{tabularx}

  \begin{tabularx}{\textwidth}{ m{2cm} c }
    3D VAE GAN
    &
    \begin{minipage}{1.0\linewidth}
      \includegraphics[width=0.8\linewidth]{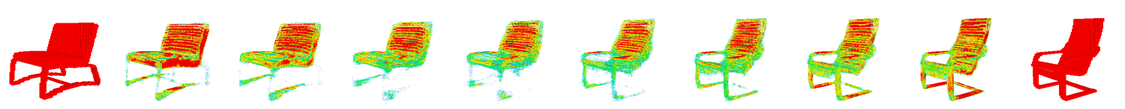}
    \end{minipage}
    \\  
    
    3D VAE 
    &
    \begin{minipage}{1.0\linewidth}
      \includegraphics[width=0.8\linewidth]{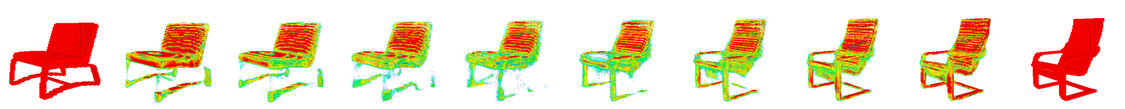}
    \end{minipage}
    \\ 
   
    Struct-Aware 
    &
    \begin{minipage}{1.0\linewidth}
      \includegraphics[width=0.8\linewidth]{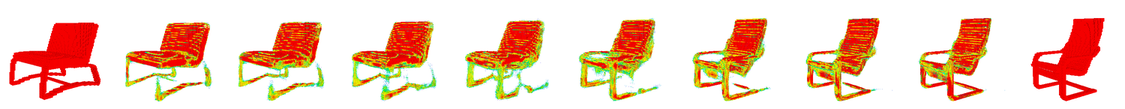}
    \end{minipage}
    \\ \hline
  \end{tabularx}
  
  \begin{tabularx}{\textwidth}{ m{2cm} c }
    3D VAE GAN
    &
    \begin{minipage}{1.0\linewidth}
      \includegraphics[width=0.8\linewidth]{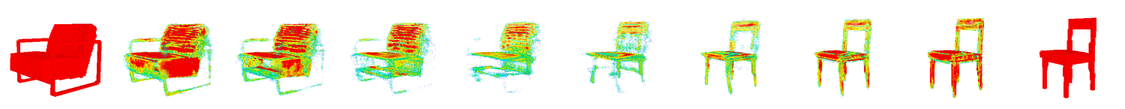}
    \end{minipage}
    \\  
    
    3D VAE 
    &
    \begin{minipage}{1.0\linewidth}
      \includegraphics[width=0.8\linewidth]{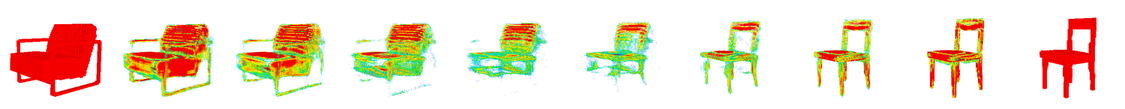}
    \end{minipage}
    \\ 
   
    Struct-Aware 
    &
    \begin{minipage}{1.0\linewidth}
      \includegraphics[width=0.8\linewidth]{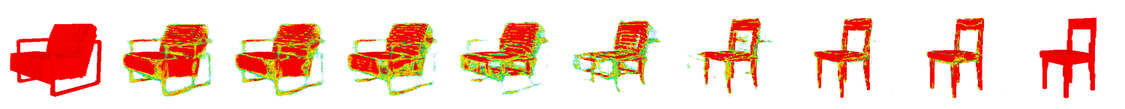}
    \end{minipage}
    \\ \hline
  \end{tabularx}
  
  \begin{tabularx}{\textwidth}{ m{2cm} c }
    3D VAE GAN
    &
    \begin{minipage}{1.0\linewidth}
      \includegraphics[width=0.8\linewidth]{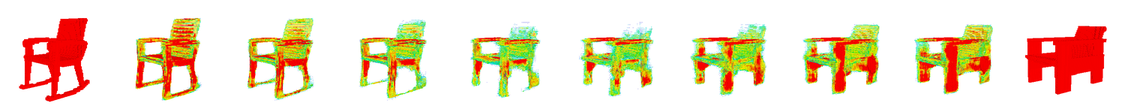}
    \end{minipage}
    \\  
    
    3D VAE 
    &
    \begin{minipage}{1.0\linewidth}
      \includegraphics[width=0.8\linewidth]{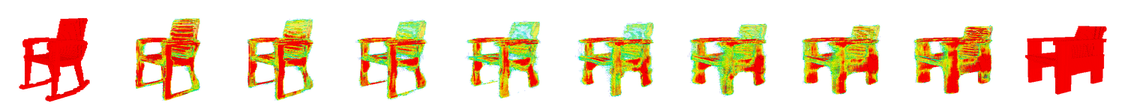}
    \end{minipage}
    \\ 
   
    Struct-Aware 
    &
    \begin{minipage}{1.0\linewidth}
      \includegraphics[width=0.8\linewidth]{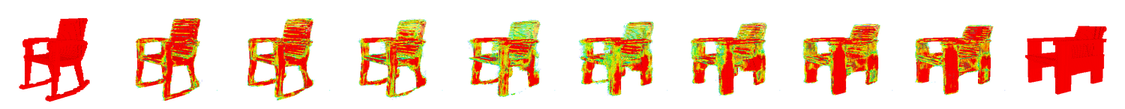}
    \end{minipage}
    \\ \hline
  \end{tabularx}

  \begin{tabularx}{\textwidth}{ m{2cm} c }
    3D VAE GAN
    &
    \begin{minipage}{1.0\linewidth}
      \includegraphics[width=0.8\linewidth]{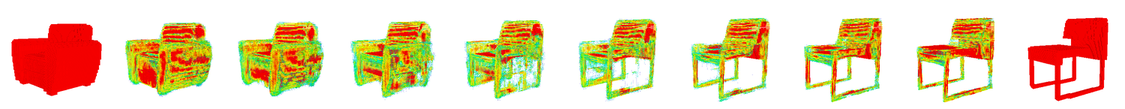}
    \end{minipage}
    \\  
    
    3D VAE 
    &
    \begin{minipage}{1.0\linewidth}
      \includegraphics[width=0.8\linewidth]{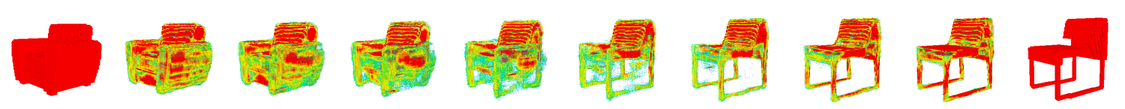}
    \end{minipage}
    \\ 
   
    Struct-Aware 
    &
    \begin{minipage}{1.0\linewidth}
      \includegraphics[width=0.8\linewidth]{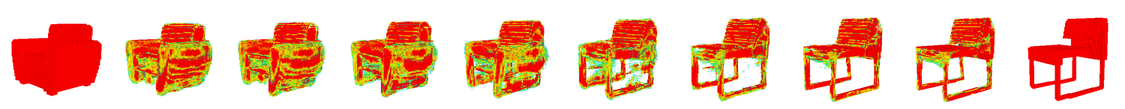}
    \end{minipage}
    \\ \hline
  \end{tabularx}

\caption{Additional comparisons between 3D VAE-GAN~\cite{wu3dGAN} (an adversarial approach), 3D VAE, and the Structure-Aware method.} \label{fig:gan_comp}
\end{figure*}

\begin{figure*}
\centering
\includegraphics[height=0.80\textheight]{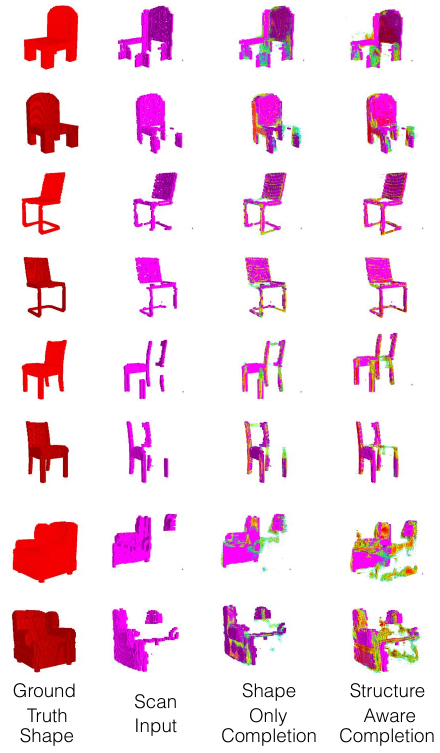}
\caption{Visualizations of additional shape completion results from two different views.} \label{fig:scan-completion}
\end{figure*}

\end{document}